\documentclass[11pt]{article}
\PassOptionsToPackage{hyphens}{url} 
\usepackage[a4paper,margin=1in]{geometry}
\usepackage[utf8]{inputenc}
\usepackage[T1]{fontenc}
\usepackage{lmodern}
\usepackage{microtype}
\usepackage{graphicx}
\usepackage{amsmath,amssymb}
\usepackage[numbers]{natbib}
\usepackage[hidelinks]{hyperref}
\usepackage{booktabs}
\usepackage{caption}
\usepackage{algorithm}
\usepackage{algorithmic}
\usepackage{xcolor}
\usepackage{xspace}
\usepackage{url}
\usepackage{authblk}

\ifx\figurename\undefined \def\figurename{Figure}\fi
\renewcommand{\figurename}{Fig.}

\newcommand{\Sect}[1]{Sec.~\ref{#1}}
\newcommand{\Fig}[1]{Fig.~\ref{#1}}
\newcommand{\Tbl}[1]{Tab.~\ref{#1}}
\newcommand{\Equ}[1]{Eq.~\ref{#1}}

\newcommand{\proj}{\textsc{PerturbMap}\xspace}
\newcommand{\projgr}{\underline{\textsc{PerturbMap}-GR}\xspace}
\newcommand{\projadap}{\underline{\textsc{PerturbMap}-ADAP}\xspace}

\newcommand{\projmax}{\underline{\textsc{PerturbMap}-MAX}\xspace}
\newcommand{\projrviii}{\underline{\textsc{PerturbMap}-R8}\xspace}
\newcommand{\projrxxxii}{\underline{\textsc{PerturbMap}-R32}\xspace}
\newcommand{\poolref}{\underline{\textsc{TokPool}}\xspace}
\newcommand{\baseLR}{\underline{\textsc{LowRank}}\xspace}

\graphicspath{{fig/}{figures/}}

\title{PerturbMap: Cross-Context Transfer of Single-Cell Perturbation Responses}

\author[1]{Panpan Cui}
\author[2]{Yiqi Liu}
\author[3]{Wenhao Sun}

\affil[1]{School of Advanced Interdisciplinary Sciences, University of Chinese Academy of Sciences}
\affil[2]{Institute of Computing Technology, CAS; University of Chinese Academy of Sciences}
\affil[3]{Hong Kong University of Science and Technology}

\date{}

\begin{document}
\maketitle



\begin{abstract}
Single-cell perturbation atlases rarely measure every intervention in every cellular context: a query perturbation is often observed in one or more source contexts but missing in the recipient context where its effect is needed.
Ignoring those measured responses discards query-specific experimental evidence, whereas copying or weakly calibrating them across contexts risks transferring the wrong signal.
We propose \proj, which predicts a missing recipient-context effect by combining a recipient-local low-rank base with accepted proposals that transport the same perturbation's measured source responses through source-to-recipient ridge experts fit on paired training perturbations, with proposal weights determined by route reliability estimated on validation anchors.
On the Perturb-CITE-seq melanoma cohort, \proj improves full-effect MSE by 4.1\% over a recipient-local low-rank base and achieves lower MSE than FedAvg, zero-response, raw-copy, calibrated-copy, and identity-shuffled affine controls. It remains within $2.82\times10^{-6}$ MSE of our centralized token-matched pooled reference, which uses a stronger training interface. A condition-mean specificity diagnostic shows the same direction: same-recipient top-10 counterpart retrieval by cosine increases from 74.5\% for the low-rank
base to 80.5\% for \proj.
\end{abstract}

\section{Introduction}
\label{sec:introduction}

Single-cell perturbation screens reveal how genetic interventions reshape transcriptional programs in specific biological contexts~\cite{replogle2022perturbseq,frangieh2021perturbcite,jiang2025pathway}.
As these screens grow, their missingness becomes more structured rather than less important. A perturbation may be measured in several source contexts but absent from the recipient context where a scientist wants to reason about pathway regulation, cancer-state specificity, or experimental prioritization. In this setting, the source response is not merely side information about the perturbation. It is an experiment on the exact query intervention, observed in a different cellular context. The scientific question is how to reuse that evidence without erasing context.

\begin{figure}[h]
\centering
\includegraphics[width=\columnwidth]{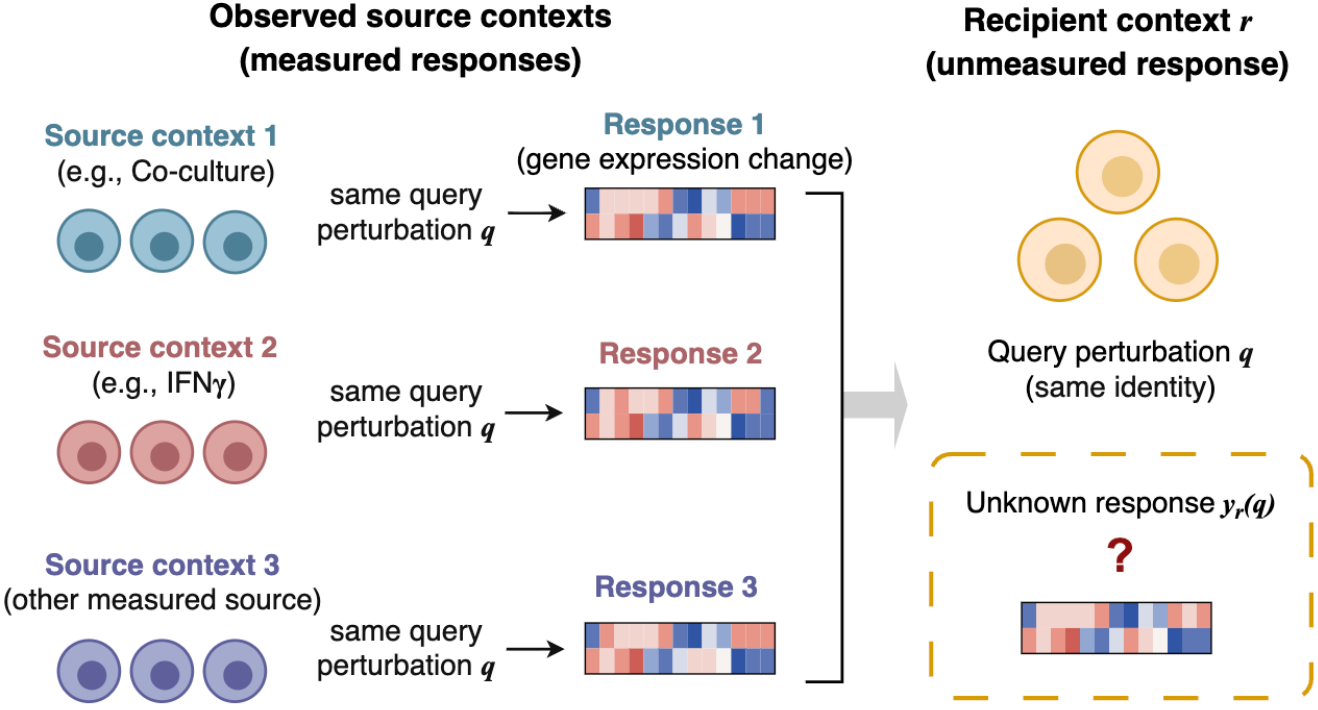}
\caption{Motivation.}
\label{fig:motivation}
\end{figure}

We call this problem source-observed, recipient-unmeasured perturbation response prediction, as shown in \Fig{fig:motivation}. For a query perturbation, model receives one or more control-relative source response vectors and must predict the hidden control-relative response in a recipient context. 

This setting differs from standard unseen perturbation prediction, where models such as scGen, GEARS and CPA generalize from perturbation and covariate representations~\cite{lotfollahi2019scgen, roohani2023gears,lotfollahi2023cpa}. 
The input is therefore richer than a perturbation identifier but still excludes the recipient outcome.

The central technical challenge is selective trust. Directly copying a source response uses the query experiment but assumes away context-specific rewiring.
A global multi-source regression can borrow information across observed contexts, but it must share one error model across perturbations whose transfer behavior may differ by gene program and recipient coordinate. 
Uniform late fusion and stacking are strong multi-view alternatives~\cite{wolpert1992stacked},
yet they treat all calibrated routes as equally reliable.
What is missing is an inference rule that learns directed source-to-recipient response maps and, using only paired training
perturbations, decides which routes should influence a sealed query.

\proj addresses this challenge with train-only reliability-weighted response transport. In the Frangieh evaluation, \proj builds a shared train-only response basis from training effects and trains a recipient-local base predictor in that basis. Each observed source context then defines a ridge expert in the shared coordinate system. Before any held recipient response is opened, a disjoint
training-anchor validation split fixes the interpolation strength and reliability of every route. During inference, the query's measured source responses are propagated through the accepted experts and combined using fixed-route scores. The result is a response-reuse rule whose degrees of freedom are set by training anchors rather than by the identities it is asked to predict.

We evaluate this idea on a manifest-backed Frangieh et al.\ Perturb-CITE-seq cohort with three cellular contexts and five identity-held outer folds~\cite{frangieh2021perturbcite}.
\proj improves full-effect MSE over the registered base predictor, FedAvg without query source-response tokens, calibrated and raw copy, and zero-response prediction, with a significant paired gain versus the base.
A centralized token-matched pooled reference remains only marginally better, so train-only reliability weighting recovers nearly the same accuracy without a pooled training interface.
\proj helps most held identities but not all; a query-adaptive harm-control gate further reduces harm directionally and is reported as an ablation.
On a secondary Jiang et al.\ multi-source Perturb-seq check, reliability weighting also yields a small paired gain over uniform fusion, without using that cohort to choose the Frangieh operating point.

This paper makes the following contributions:

\begin{itemize}
    \item We define source-observed, recipient-unmeasured perturbation response prediction as an identity-held test of whether measured source effects can
    help fill missing recipient effects.
    \item We introduce \proj, a train-only reliability-weighted transport method
    that maps source response tokens into recipient coordinates and combines
    accepted routes using validation-residual reliability.
    \item We evaluate a query-adaptive harm-control gate as an ablation, finding
    directional MSE and harm-rate improvements that require stronger validation
    before becoming a headline contribution.
    \item We evaluate \proj on a manifest-backed multi-context
    Perturb-CITE-seq cohort with information-matched source-token controls,
    identity shuffles, paired uncertainty estimates, secondary response-shape
    diagnostics, and explicit harm-rate reporting.
\end{itemize}

\section{Related Work}
\label{sec:related-work}

\paragraph{Single-cell perturbation prediction.}
Perturbation models such as GEARS and CPA predict responses for unseen
interventions, combinations, or covariate settings by generalizing from
perturbation, covariate, and basal-state
representations~\cite{roohani2023gears,lotfollahi2023cpa}. Benchmarks further
show that perturbation-response prediction contains distinct information
regimes~\cite{wu2024perturbench,wei2026benchmarking}. \proj studies a
source-observed regime: the query perturbation has measured responses in source
contexts, and the task is to reuse them to predict an unmeasured
recipient-context effect.

\paragraph{Context transfer and virtual-cell models.}
Context-transfer and virtual-cell methods model shifts across conditions,
including neural optimal transport, conditional transport, and broader
virtual-cell predictors~\cite{bunne2023not,driessen2025cmonge,yu2026scdfm,
chi2025departures,fu2026strand,chen2026scale,jiang2026ocoot}. These methods
support the premise that context variation is structured, but usually align
full cell populations or generate cell-level counterfactuals from descriptors,
controls, or pretraining corpora. \proj instead transports condition-level
effects for the same query intervention while keeping the recipient outcome
sealed.

\paragraph{Multi-source adaptation and reliability-aware fusion.}
Federated optimization, domain adaptation, and federated personalization address
heterogeneous sources through distributed training or source-invariant/source-aware
predictors~\cite{mcmahan2017fedavg,sahu2020fedprox,ganin2016dann,
sun2016coral,li2021fedbn,kairouz2021advances}. Multi-view fusion methods such
as averaging or stacking can combine source predictions, but can be brittle
when reliability varies by directed source-recipient route. \proj instead
uses measured source responses as query-time tokens and calibrates each route
on paired training anchors to form a train-only reliability-weighted consensus.
\section{Problem Setup}
\label{sec:problem-setup}

\paragraph{Condition-level perturbation response.}
Let $c\in\mathcal{C}$ denote a cellular context and let $p$ denote a genetic
perturbation. On a fixed gene axis of dimension $D$, the context-local
control-relative response is
\begin{equation}
    y_c(p)=
    \mathbb{E}[X\mid p,c]-\mathbb{E}[X\mid \mathrm{control},c]
    \in\mathbb{R}^{D}.
    \label{eq:condition-effect}
\end{equation}
In the Frangieh et al. evaluation, the empirical response uses the condition
mean minus a context-local control mean on the fixed 5{,}000-gene axis.
The prediction target is the full control-relative effect vector, not a cell-level
distribution or a perturbation label.
This condition-level contract is intentionally narrower
than distributional counterfactual generation: it asks whether a missing mean perturbation effect can be completed in an atlas, not whether the full within-condition
single-cell state distribution is reproduced.

\paragraph{Source-observed, recipient-unmeasured queries.}
For a recipient context $r$, let
$\mathcal{S}_r\subseteq\mathcal{C}\setminus\{r\}$ denote the observed source contexts; in the Frangieh evaluation, these are the non-recipient contexts.
The training identities $\mathcal{I}_{\mathrm{train}}$ have paired responses
in the source contexts and in the recipient context.
For a held query perturbation
$q$, the available query input is
\begin{equation}
    \begin{aligned}
    \mathcal{X}_r(q)=\Bigl(&
    \{y_s(q)\}_{s\in\mathcal{S}_r},\ r,\\
    &\{(y_s(p),y_r(p))\}_{\substack{
    p\in\mathcal{I}_{\mathrm{train}}\\
    s\in\mathcal{S}_r}}
    \Bigr).
    \end{aligned}
    \label{eq:available-input}
\end{equation}
The held recipient response $y_r(q)$ is unavailable until after a sealed
prediction has been written. The model output is
\begin{equation}
    \widehat y_r(q)=
    F_r\left(\{y_s(q)\}_{s\in\mathcal{S}_r};
    \mathcal{I}_{\mathrm{train}}\right).
    \label{eq:transport-task}
\end{equation}
This contract directly tests whether measured source responses of the same
perturbation can improve recipient prediction 
beyond source copying, zero-response nulls, and information-matched fusion baselines.

\paragraph{Identity-held information boundary.}
All evaluation splits are by perturbation identity.
For each recipient-by-fold prediction,
the held identities are excluded from recipient coordinate construction,
source-to-recipient map fitting, train-only reliability calibration, and any
baseline training. Source responses for held identities remain available only
because they are the query measurements in \Equ{eq:transport-task}. The
scoring program authenticates every sealed prediction artifact before opening
any held recipient matrix rows and persists per-condition scores rather than
held response vectors.

\section{Method}
\label{sec:method}

\subsection{Overview}

\begin{figure*}[t]
\centering
\includegraphics[width=\textwidth]{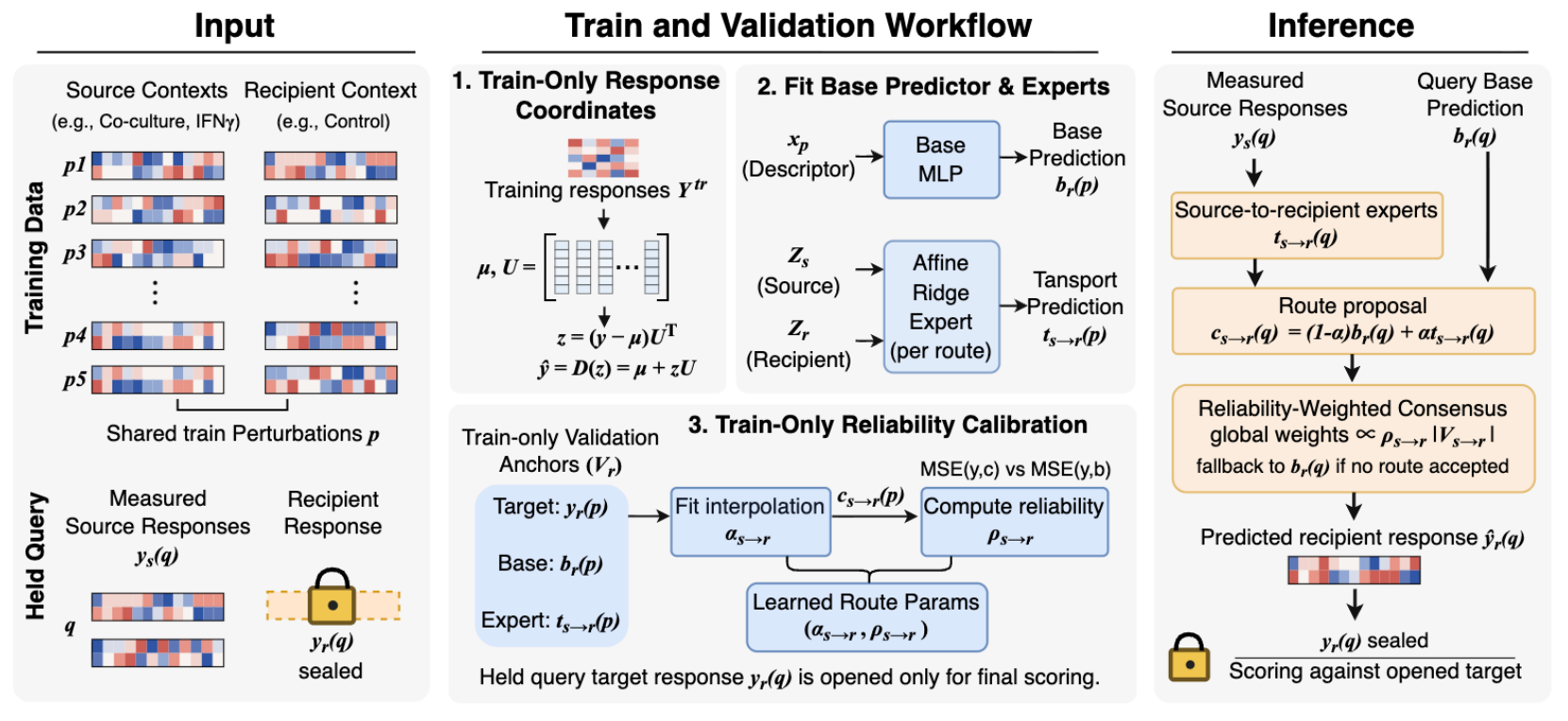}
\caption{\proj protocol overview.}
\label{fig:protocol}
\end{figure*}

\Fig{fig:protocol} summarizes the \proj workflow.  \proj predicts a missing recipient response by combining a recipient-local base
predictor with source-specific response-transport proposals. The method has
four modules. First, it builds train-only response coordinates. Second, it fits a recipient-local low-rank base predictor that does not use query
source-response tokens.
Third, it fits one regularized source-to-recipient expert per observed
source context. Fourth, it uses a disjoint training-anchor validation split to
estimate each route's interpolation strength and global route reliability before
predicting held queries. We additionally evaluate a low-capacity query-adaptive
gate as a harm-control component.

\subsection{Train-Only Response Coordinates}

Let $Y^{\mathrm{tr}}$ denote the collection of training response rows available
across clients. In the Frangieh implementation, \proj fits a shared rank-$K$
train-only covariance-sketch basis with QR iterations,
\begin{equation}
    z_c(p)=(y_c(p)-\mu)U^\top,\qquad
    \mathcal{D}(z)=\mu+zU,
    \label{eq:response-coordinate}
\end{equation}
where $\mu$ and $U$ are computed without held recipient rows. This shared basis is the coordinate system used by the recipient-local base
predictor, the source response tokens, and the source-to-recipient route maps.
Decoded predictions are
evaluated on the original full-gene axis, so the low-rank representation is an
estimation device rather than a reduced evaluation target.

\subsection{Recipient-Local Base Predictor}

The base predictor $b_r(p)$ is the model \proj falls back to when source
transport is not validated. It predicts response-coordinate coefficients from
the perturbation descriptor using a client-local network: LayerNorm, a
width-128 linear layer, GELU, and a linear rank-$K$ output head. The decoded
prediction is $\mu+g_r(x_p)U$, where $x_p$ is the fixed perturbation descriptor.
The network is trained on recipient-train identities with coefficient-space MSE;
round selection uses decoded full-effect validation MSE. In Frangieh, $x_p$ is
not a one-hot perturbation identity. It is
the targeted gene's train-only control descriptor: correlations with the 64 most
variable control-anchor genes, plus control mean, standard deviation, and
detection frequency. Thus the base has no query-specific perturbation-effect
token; the measured source response is the only query-visible effect
measurement. The Frangieh run uses rank $K=16$, batch size 16, learning rate
$10^{-3}$, weight decay $10^{-4}$, three local epochs per round, at most 100
rounds, and validation early stopping with patience 15.

\subsection{Source-to-Recipient Experts}


For a recipient $r$ and source $s\in\mathcal{S}_r$, the paired training
perturbations define source and recipient coordinate matrices $Z_s$ and $Z_r$
in the shared train-only coordinate system. \proj fits an affine ridge response
map by centering the paired coordinates. Let $\widetilde Z_s$ and
$\widetilde Z_r$ denote the centered source and recipient matrices. The linear
map is
\begin{equation}
    A_{s\to r} =
    \arg\min_A
    \lVert \widetilde Z_s A-\widetilde Z_r\rVert_F^2
    + \lambda \eta_s \lVert A\rVert_F^2,
    \label{eq:source-expert}
\end{equation}
where $\eta_s$ is the average source-coordinate energy used to scale the ridge
penalty; the affine bias is the difference between the recipient anchor mean
and the mapped source anchor mean. Given a held query source response $y_s(q)$,
the uncalibrated expert prediction $t_{s\to r}(q)$ is obtained by applying this
centered affine map and decoding back to the full-gene axis.
This expert uses the actual observed response of the query perturbation
in source context $s$, but it never reads $y_r(q)$.

\subsection{Train-Only Reliability Calibration}

Let $b_r(p)$ be the recipient-local low-rank base prediction for perturbation
$p$ and let $t_{s\to r}(p)$ be the decoded source-transport prediction from
source $s$ to recipient $r$. On disjoint validation anchors
$\mathcal{V}_r$, \proj fits a route-specific interpolation coefficient
\begin{equation}
    \alpha_{s\to r} =
    \operatorname{clip}_{[0,1]}
    \frac{\sum_{p\in\mathcal{V}_r}
    \langle y_r(p)-b_r(p),\,t_{s\to r}(p)-b_r(p)\rangle}
    {\sum_{p\in\mathcal{V}_r}
    \lVert t_{s\to r}(p)-b_r(p)\rVert_2^2}.
    \label{eq:route-alpha}
\end{equation}
The route proposal is
\begin{equation}
    c_{s\to r}(q) =
    (1-\alpha_{s\to r})b_r(q)+\alpha_{s\to r}t_{s\to r}(q).
    \label{eq:route-proposal}
\end{equation}
Its train-only route score is the relative validation-MSE reduction
\begin{equation}
    \rho_{s\to r} =
    \max\left(0,\,
    1-
    \frac{\operatorname{MSE}_{p\in\mathcal{V}_r}(y_r(p),c_{s\to r}(p))}
    {\operatorname{MSE}_{p\in\mathcal{V}_r}(y_r(p),b_r(p))+\epsilon}
    \right),
    \label{eq:route-confidence}
\end{equation}
with $\epsilon=10^{-12}$ in the Frangieh implementation. This score is a
validation-split route statistic, not a query-level uncertainty bound.

\paragraph{Query-adaptive harm control.}
\proj also evaluates a second, low-capacity query-route gate on the same validation
anchors. For each validation query $p$ and candidate route $s\to r$, we build a
label-free feature vector containing: query distance to recipient-train
perturbation descriptors; source-response distance to source-train response
anchors; source-to-base displacement; source-response norm and sparsity;
candidate disagreement across source experts; train-anchor route residual MSE;
and the route's $\alpha_{s\to r}$, $\rho_{s\to r}$, and validation support. A
regularized logistic model estimates

\begin{equation}
\begin{aligned}
\pi_{s\to r}(q)
&=
P\Big[
\operatorname{MSE}(c_{s\to r}(q),y_r(q))\\
&\quad <
\operatorname{MSE}(b_r(q),y_r(q))
\mid \phi_{s\to r}(q)
\Big]
\end{aligned}
\end{equation}

using validation-anchor labels only. The harm-control query weight is
\begin{equation}
    w_{s\to r}(q)=\rho_{s\to r}\pi_{s\to r}(q).
\end{equation}
The main Frangieh configuration, denoted \projgr, uses the global
route-reliability aggregate
\begin{equation}
    \widehat y_r(q)=
    \sum_{s\in\mathcal{A}^{\mathrm{gr}}_r(q)}
    \frac{\rho_{s\to r}|\mathcal{V}_{s\to r}|}
    {\sum_{s'\in\mathcal{A}^{\mathrm{gr}}_r(q)}
    \rho_{s'\to r}|\mathcal{V}_{s'\to r}|}
    c_{s\to r}(q),
    \label{eq:global-reliability-prediction}
\end{equation}
where $\mathcal{A}^{\mathrm{gr}}_r(q)$ is the set of query-available source
routes with positive validation reliability, and
$\mathcal{V}_{s\to r}\subseteq\mathcal{V}_r$ denotes the validation anchors
available for route $s\to r$. The predictor falls back to $b_r(q)$ if no source
route is accepted.

The query-adaptive harm-control ablation, denoted \projadap, instead uses
\begin{equation}
    \widehat y_r(q)=
    \sum_{s\in\mathcal{A}^{\mathrm{adap}}_r(q)}
    \frac{w_{s\to r}(q)|\mathcal{V}_{s\to r}|}
    {\sum_{s'\in\mathcal{A}^{\mathrm{adap}}_r(q)}
    w_{s'\to r}(q)|\mathcal{V}_{s'\to r}|}
    c_{s\to r}(q),
    \label{eq:query-adaptive-prediction}
\end{equation}
where $\mathcal{A}^{\mathrm{adap}}_r(q)$ is the set of query-available source
routes with positive validation reliability and query probability above a
validation-selected threshold. The predictor falls back to $b_r(q)$ if no source
route is accepted. We report \projadap as a directional harm-control ablation
because its Frangieh MSE confidence interval versus \projgr crosses zero.

\subsection{Split Discipline}

For each Frangieh outer fold, 40 perturbation identities are held out for
testing. The 200 supported identities each serve as a held recipient query
exactly once in the frozen Frangieh protocol. Recipient counts are 67, 67, and
66 across the three contexts, and the non-recipient contexts provide
query-visible source-response tokens. The
remaining identities are split into train and a disjoint inner-validation set
using the frozen split seed 20{,}260{,}718 and validation fraction 0.2. The
outer-test recipient response is never used to fit the base
predictor, response coordinates, ridge maps, $\alpha$, $\rho$, optional query
gate, route admission, or source aggregation. Inner-validation identities are
used for early stopping, ridge selection, and the reliability quantities in
\Equ{eq:route-alpha}--\Equ{eq:query-adaptive-prediction}. All
non-recipient contexts for held or validation identities remain available only
as source-response tokens.

\subsection{Frozen Outer-Fold Inference}

All hyperparameters, folds, source contexts, and comparison methods are frozen
before held-test scoring. Each outer-fold job writes a result artifact and a
manifest that records the code revision, source-file hashes, input checksum,
fold, held identities, and read audit. The scoring program authenticates the
manifest before aggregating held recipient effects. This execution boundary is
part of the method because it prevents reliability estimates, coordinate
choices, ridge selection, or baseline selection from using the target responses
they are meant to predict.

\section{Experiments}
\label{sec:experiments}

\subsection{Frangieh et al. Multi-Context Evaluation}

The primary evaluation uses the Frangieh et al. multimodal Perturb-CITE-seq
melanoma immune-evasion cohort~\cite{frangieh2021perturbcite}. The processed
manifest-backed matrix contains 109{,}155 cells, 5{,}000 genes, and three selected cellular contexts: Co-culture, Control, and IFN$\gamma$. The frozen
support contains 200 shared perturbation identities.

\paragraph{Identity-held query protocol.}
Each of the 200 identities is scored exactly once as a held recipient query,
with recipient assignment balanced as 67, 67, and 66 across contexts. For a
query assigned to recipient $r$, the two non-recipient contexts provide
query-visible source-response tokens, while $y_r(q)$ remains sealed until
after prediction. The primary score therefore aggregates 200
source-observed, recipient-unmeasured queries rather than a full
recipient-by-identity Cartesian product.

The evaluation uses five identity-held outer folds of 40 test identities
each. Within each fold, the remaining 160 identities are split into train and
a disjoint inner-validation set with frozen seed 20{,}260{,}718 and
validation fraction 0.2. Fitting and selection use only train or validation
identities under the split discipline of \Sect{sec:method}; held recipient
responses are never read. Non-recipient contexts of held or validation
identities remain available only as source-response tokens. The run freezes
the data artifacts, contexts, identities, folds, rank-16 response coordinates, ridge
grid $\{10^{-3},10^{-2},10^{-1},1,10\}$, and \projgr aggregation; ridge is
chosen by inner-validation MSE, and held effects are scored once.

\paragraph{Metrics and uncertainty.}
The primary metric is condition-level full-effect MSE on the fixed
5{,}000-gene axis. Secondary metrics are top-20 absolute-effect MSE,
effect-vector Pearson correlation, and cosine similarity, each computed per
identity and then averaged; top-20 genes are selected from the absolute true
recipient effect. A win (harm) is a held identity with strictly lower
(higher) full-effect MSE than a stated comparator; ties count as neither.
On Frangieh the default comparator is \baseLR; on Jiang it is uniform expert
fusion. 
We also report top-20 gene overlap, sign agreement on the true
top-20 genes, cosine-based top-10 retrieval relative to \baseLR, average
per-fold communication payload, and five-fold runner wall time.
Uncertainty uses outer-fold-stratified paired bootstrap over perturbation
identities; we report 95\% confidence intervals for selected
method-minus-comparator MSE contrasts, and treat aggregate gaps without
intervals as descriptive only.

\subsection{Baselines, References, and Ablations}

Comparisons are organized by the claim they test.

\paragraph{Does the query source response add information?}
\baseLR is the registered recipient-local LowRank predictor sharing the
response basis and optimization budget of \proj but receiving no query
source-response token.
FedAvg is a no-token federated MLP reference outside the low-rank
response-coordinate architecture. Zero response is the low-variance null.

\paragraph{Is the gain due to identity-aligned transport?}
Raw copy directly transfers source full-gene effects through the same route-admission and aggregation rule as \projgr. 
Calibrated copy replaces the fitted affine map with the identity map in
response coordinates while retaining the validation-calibrated interpolation,
route admission, and aggregation. 
Identity-shuffled affine transport keeps the model class and
anchor count but applies a route-specific derangement of source--recipient
anchor pairings. Two equal-token adapted baselines receive the same
query-visible source tokens without directed transport: source-token ridge
and a validation-selected two-layer source-token MLP.

\paragraph{How close is client-local training to a stronger interface?}
\poolref matches the query-time source token, train-only basis, affine
routes, route admission, and source-token policy of \proj, but replaces client-local
recipient training with a centrally shared trunk and one linear rank-$K$
head per recipient. It is a performance target rather than an
interface-matched baseline; Frangieh tables report its absolute MSE and gap
to \projgr, not a $\Delta$ against \baseLR.

\paragraph{Which components and operating points matter?}
To test whether \baseLR is underpowered, we evaluate an RBF kernel low-rank
regressor and a validation-selected blend with \baseLR. We also probe
reliability-modulated uncertainty, validation-selected $\alpha$ bounds from
$\{[0,1],[-0.25,1.25],[-0.5,1.5],[-1,2]\}$, and directed routes capped at
32, 64, or 96 paired train anchors. Reported Frangieh variants are
\projgr, \projadap, \projmax, \projrviii, and \projrxxxii: \projgr uses
route-level $\rho_{s\to r}$; \projadap multiplies by the query gate;
\projmax selects the accepted route with largest $\rho$; and the rank
variants change only $K$. Uniform fusion, no-$\alpha$, and no-$\rho$ are treated as diagnostic
operating-point probes, while the Jiang cohort provides the secondary sealed
reliability-fusion comparison under richer source support.

\paragraph{Secondary reliability-fusion check.}
We report a previously sealed Jiang et al. multi-source Perturb-seq
check~\cite{jiang2025pathway} only to isolate reliability weighting under
richer source support. It is an independent frozen protocol: six recipient
contexts, five observed sources per recipient, rank-64 coordinates, ridge
1.0, and cross-fitted posterior-consensus reliability. Comparators are
uniform expert averaging, out-of-fold expert stacking, and a joint
multi-source affine ridge. We denote the Jiang reliability-weighted entry by
\proj; it is not Frangieh \projgr. Harm and win rates are relative to
uniform fusion.

\subsection{Development Boundary}

Earlier Replogle and SciPlex3 analyses were used only for method development
and successor search. They motivate the current response-reuse formulation,
but they are not final test evidence for the Frangieh claims in this paper.

\section{Results}
\label{sec:results}

\subsection{Primary Frangieh Results}

Across 200 identity-held perturbations, measured source responses improve
recipient prediction beyond \baseLR. \Tbl{tab:frangieh-main} shows that
\projgr reduces full-effect MSE by $6.80\times10^{-5}$, or 4.1\%. The paired
95\% confidence interval is $[-1.16,-0.33]\times10^{-4}$ and excludes zero.
\projgr improves 161 of 200 identities and harms 39 relative to \baseLR. Its
aggregate MSE is also $3.66\times10^{-4}$ to $3.81\times10^{-4}$ below the two
no-token controls and $5.6\times10^{-5}$ to $6.3\times10^{-5}$ below the
copy and shuffled-affine controls; the shuffled contrast is paired-significant
in \Tbl{tab:reviewer-diagnostics}, while the copy gaps are reported as
aggregate MSE differences. These comparisons support the main claim that
identity-aligned source responses provide predictive information beyond
recipient-only behavior, direct transfer, and marginal calibration. The
remaining gap to the stronger centralized \poolref is $2.82\times10^{-6}$.
\projadap reduces harm to 32 identities, but its paired confidence interval
against \projgr includes zero.

\begin{table}[!htbp]
\centering
\setlength{\tabcolsep}{3pt}
\renewcommand{\arraystretch}{1.08}
\begin{tabular}{lcc}
\toprule
Method &
\shortstack{MSE\\($\times10^{-3}$)} &
\shortstack{$\Delta$ vs.\ \baseLR\\($\times10^{-5}$)} \\
\midrule
\textbf{\projgr} & \textbf{1.581} & $\mathbf{-6.80}$ \\
\projadap & 1.569 & $-8.04$ \\
\poolref & 1.578 & -- \\
\baseLR & 1.649 & $0$ \\
Calibrated copy & 1.644 & $-0.55$ \\
Raw copy & 1.640 & $-0.86$ \\
Shuffled affine & 1.637 & $-1.20$ \\
FedAvg & 1.947 & $+29.8$ \\
Zero response & 1.961 & $+31.2$ \\
\bottomrule
\end{tabular}
\caption{Frangieh et al. multi-context evaluation.}
\label{tab:frangieh-main}
\end{table}

\begin{figure}[h]
\centering
\includegraphics[width=\linewidth]{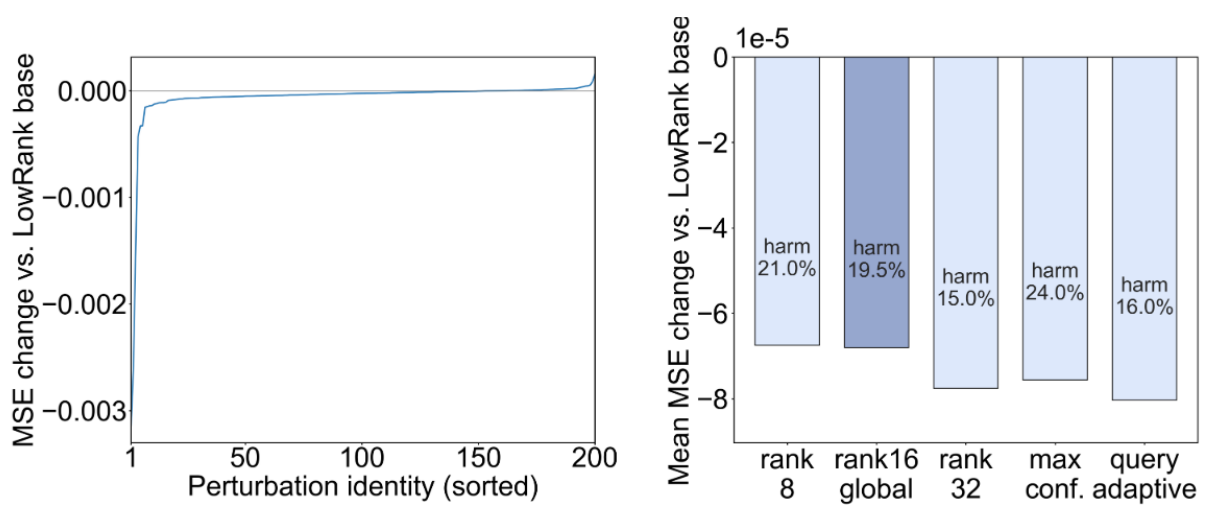}
\caption{Identity-level gains and operating-point sensitivity.}
\label{fig:frangieh-results}
\end{figure}

\Fig{fig:frangieh-results} summarizes identity-level gains and operating-point
sensitivity on Frangieh. The aggregate gain is broad but not uniform across
identities, and the ablation panel shows that nearby operating-point
differences are much smaller than the gain over \baseLR.
\Tbl{tab:frangieh-all-metrics} tests whether the improvement extends beyond
full-effect MSE.
\projgr reduces top-20 MSE by about 12\% against direct-copy
controls and by 28--35\% against the no-token controls. Pearson and cosine
increase by 2.4--2.8 percentage points against direct copy. Harm falls to
19.5\%, compared with 49.5\% for raw copy and 54.5\% for calibrated copy. 
Thus, source-response transport improves both effect magnitude and response shape while reducing negative transfer relative to copy controls.
Additional diagnostics agree:
top-20 gene overlap rises from 9.23\% to 10.63\%, direction agreement rises
from 89.73\% to 90.75\%, and cosine-based top-10 retrieval rises from 74.5\%
to 80.5\% relative to \baseLR.


\begin{table}[!htbp]
\centering
\setlength{\tabcolsep}{3pt}
\begin{tabular}{lcccc}
\toprule
Method & Top-20 ($\times10^{-2}$) & Pear. & Cos. & Harm \\
\midrule
\projgr & 2.85 & 0.448 & 0.450 & 0.195 \\
\projadap & 2.72 & 0.449 & 0.451 & 0.160 \\
\poolref & 2.83 & 0.448 & 0.451 & 0.220 \\
Calibrated copy & 3.27 & 0.420 & 0.422 & 0.545 \\
Raw copy & 3.23 & 0.424 & 0.426 & 0.495 \\
Shuffled affine & 3.32 & 0.432 & 0.435 & 0.220 \\
FedAvg & 3.93 & 0.191 & 0.191 & -- \\
Zero response & 4.36 & 0.000 & 0.000 & -- \\
\bottomrule
\end{tabular}
\caption{Frangieh aggregate metrics.}
\label{tab:frangieh-all-metrics}
\end{table}

\subsection{Method Diagnostics and Operating Point}

\Tbl{tab:reviewer-diagnostics} probes alternative explanations for the main
performance gain. \projgr improves over nonlinear recipient-only baselines by
$5.46\times10^{-5}$ to $5.72\times10^{-5}$ MSE and over equal-token ridge/MLP
predictors by $3.51\times10^{-5}$ to $4.00\times10^{-5}$, winning 136--144 of
200 paired identities. The identity-shuffled control is worse by
$5.60\times10^{-5}$, isolating aligned source-to-recipient pairing. Route
support also matters: caps of 32 and 64 anchors significantly degrade MSE.

Several variants are statistically tied with \projgr (CI includes zero):
\projadap\ vs.\ \projgr ($-1.23\times10^{-5}$), uncertainty modulation
($-0.006\times10^{-5}$), $\alpha$-extrapolation ($-0.020\times10^{-5}$), the
96-anchor cap ($-0.020\times10^{-5}$), and raw-copy vs. shuffled
($+0.35\times10^{-5}$).

\begin{table}[!t]
\centering
\setlength{\tabcolsep}{4pt}
\renewcommand{\arraystretch}{1.05}
\begin{tabular*}{\columnwidth}{@{\extracolsep{\fill}}lcc@{}}
\toprule
Comparator & $\Delta$MSE ($\times10^{-5}$) & Wins / 200 \\
\midrule
\multicolumn{3}{@{}l}{\textit{All rows: $\projgr$ vs.\ comparator; CI excludes zero.}} \\
RBF kernel base      & $-5.46$ & 136 \\
Adapt. MLP/kernel    & $-5.72$ & 144 \\
Token ridge          & $-3.51$ & 138 \\
Token MLP            & $-4.00$ & 142 \\
Cap-32 anchors       & $-3.78$ & 138 \\
Cap-64 anchors       & $-1.66$ & 134 \\
Shuffled affine      & $-5.60$ & 128 \\
\bottomrule
\end{tabular*}
\caption{Frangieh method diagnostics.}
\label{tab:reviewer-diagnostics}
\end{table}


\Tbl{tab:frangieh-ablations} measures sensitivity to aggregation and response
rank. All three variants remain within $7.5\times10^{-6}$ MSE of \projgr,
which is more than $9\times$ smaller than the main gain over \baseLR.
Increasing rank from \projrviii to \projrxxxii raises communication from
4.02 to 9.96~MB without a monotonic accuracy gain. These results support
robustness to the tested operating points, not superiority of rank 16.
Accordingly, the frozen rank-16 configuration remains the headline rather
than a post-hoc best variant. It communicates 6.00~MB per fold, and the five
folds require 76.4~s of recorded runner wall time.

\begin{table}[!htbp]
\centering
\setlength{\tabcolsep}{4pt}
\begin{tabular}{lccc}
\toprule
Setting &
\shortstack{MSE\\($\times10^{-3}$)} &
\shortstack{$\Delta$MSE\\($\times10^{-5}$)} &
MB \\
\midrule
MAX & 1.573 & $-0.75$ & 6.00 \\
R32 & 1.580 & $-0.14$ & 9.96 \\
R8  & 1.575 & $-0.55$ & 4.02 \\
\bottomrule
\end{tabular}
\caption{Frangieh ablation summary for PerturbMap variants. $\Delta$MSE is measured relative to \projgr.}
\label{tab:frangieh-ablations}
\end{table}

\subsection{Robustness Across Folds and Contexts}

\Tbl{tab:frangieh-fold-mse} tests whether the aggregate result depends on one outer fold. In every fold, \projgr remains within $1.37\times10^{-5}$ MSE of \poolref. FedAvg and zero response remain worse by $3.05\times10^{-4}$ to $4.47\times10^{-4}$ across the same folds. The main separation therefore persists across held identity partitions rather than a single favorable split.

\begin{table}[!htbp]
\centering
\setlength{\tabcolsep}{3pt}
\begin{tabular}{lccccc}
\toprule
Method & F0 ($\times10^{-3}$) & F1 & F2 & F3 & F4 \\
\midrule
\projgr & 1.511 & 1.572 & 1.564 & 1.597 & 1.661 \\
\poolref & 1.511 & 1.571 & 1.564 & 1.598 & 1.647 \\
FedAvg & 1.866 & 1.917 & 1.869 & 2.012 & 2.069 \\
Zero response & 1.891 & 1.910 & 1.886 & 2.044 & 2.077 \\
\bottomrule
\end{tabular}
\caption{Frangieh outer-fold MSE.}
\label{tab:frangieh-fold-mse}
\end{table}

\Tbl{tab:frangieh-client-metrics} defines the context boundary of the aggregate
claim. Control is the hardest recipient, with MSE $1.86\times10^{-3}$, which is
30\% above Co-culture. Top-20 MSE varies by only $6.0\times10^{-4}$ across recipients, while IFN$\gamma$ has the highest Pearson and cosine values.
The principal context variation is therefore in full-effect magnitude, not in response-shape ordering.

\begin{table}[!htbp]
\centering
\setlength{\tabcolsep}{3pt}
\begin{tabular}{lcccc}
\toprule
Recipient & MSE ($\times10^{-3}$) & Top-20 ($\times10^{-2}$) & Pear. & Cos. \\
\midrule
Co-culture & 1.426 & 2.85 & 0.443 & 0.441 \\
Control & 1.857 & 2.89 & 0.445 & 0.444 \\
IFN$\gamma$ & 1.463 & 2.83 & 0.455 & 0.466 \\
\bottomrule
\end{tabular}
\caption{\projgr recipient-context metrics.}
\label{tab:frangieh-client-metrics}
\end{table}

\Tbl{tab:frangieh-routes} shows that source routes are not exchangeable under
the same split protocol. Validation confidence ranges from 0.015 to 0.093, a
$6.2\times$ difference, and transport reduces validation MSE on all six routes by roughly
$1.0\times10^{-5}$ to $1.4\times10^{-4}$.
The strongest route is IFN$\gamma$$\rightarrow$Co-culture, while routes into Control retain validation MSE near $2.1\times10^{-3}$.  This motivates route-specific trust scores on Frangieh, while the independent Jiang check separately tests reliability weighting against uniform fusion under richer source.

\begin{table}[!htbp]
\centering
\setlength{\tabcolsep}{3pt}
\begin{tabular}{lcccc}
\toprule
Route & $\alpha$ & \multicolumn{2}{c}{Val.\ MSE ($\times10^{-3}$)} & Conf. \\
\cmidrule(lr){3-4}
& & Base & Transport & \\
\midrule
Co-culture$\rightarrow$Control & 0.75 & 2.17 & 2.14 & 0.015 \\
Co-culture$\rightarrow$IFN$\gamma$ & 0.75 & 1.50 & 1.36 & 0.063 \\
Control$\rightarrow$Co-culture & 0.58 & 1.24 & 1.23 & 0.015 \\
Control$\rightarrow$IFN$\gamma$ & 0.77 & 1.50 & 1.47 & 0.019 \\
IFN$\gamma$$\rightarrow$Co-culture & 0.90 & 1.24 & 1.11 & 0.093 \\
IFN$\gamma$$\rightarrow$Control & 0.97 & 2.17 & 2.11 & 0.028 \\
\bottomrule
\end{tabular}
\caption{\projgr source-to-recipient route.}
\label{tab:frangieh-routes}
\end{table}

\subsection{Secondary Multi-Source Check}

\begin{figure}[h]
\centering
\includegraphics[width=0.8\columnwidth]{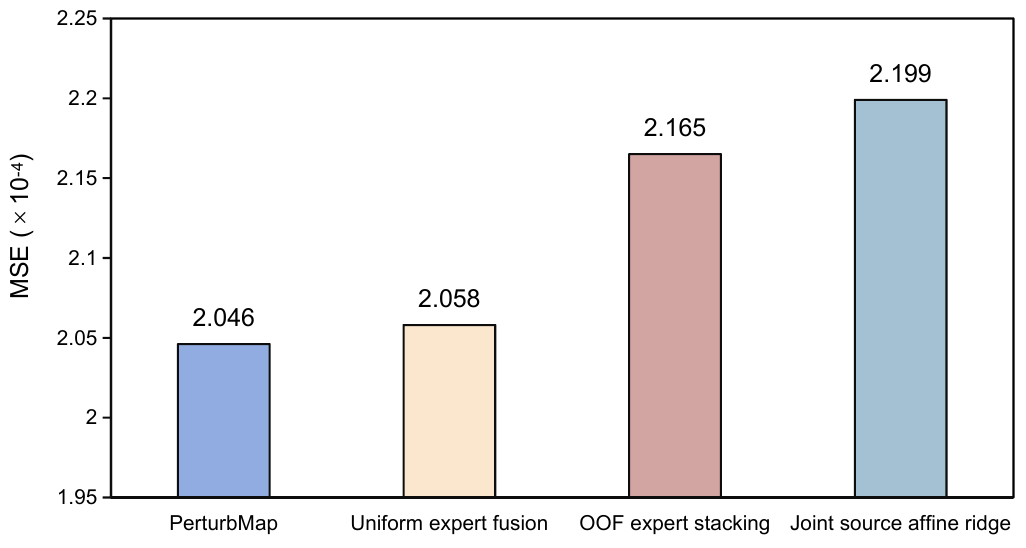}
\caption{Secondary Jiang et al. reliability-fusion check.}
\label{fig:jiang-reliability-check}
\end{figure}

\Fig{fig:jiang-reliability-check} isolates reliability weighting in a richer
five-source setting under the sealed Jiang protocol rather than Frangieh
\projgr. Reliability-weighted \proj improves over uniform fusion by
$1.23\times10^{-6}$ MSE. Its paired 95\% confidence interval is
$[-2.44,-0.30]\times10^{-6}$ and excludes zero. It also improves over
out-of-fold stacking by $1.19\times10^{-5}$. The win rate against uniform
fusion is 63.9\%, but the 36.1\% harm rate relative to uniform fusion rules
out a route-universal claim. Jiang therefore provides an independent, sealed reliability-fusion check in a richer setting.


\section{Discussion}
\label{sec:discussion}

\paragraph{Claim and limits.}
The Frangieh evaluation shows that measured source responses can serve as
query-time evidence for a missing recipient mean effect. Under a client-local,
identity-held interface, \projgr improves over \baseLR, no-token baselines, and
copy or shuffled-affine controls, while remaining close to the stronger
centralized \poolref. The claim is source-response reuse under a sealed
information boundary, not state-of-the-art virtual-cell prediction or absence
of negative transfer: \projgr still harms 19.5\% of identities relative to
\baseLR.

\paragraph{Scope and implications.}
Frangieh most directly supports identity-aligned response transport: copy and
shuffled-affine controls separate marginal calibration from paired
source-response reuse, and stronger recipient-only or equal-token predictors do
not recover the gain. Route heterogeneity motivates source-specific trust, with
Jiang providing a secondary reliability-weighted fusion check under richer
source support. The evaluation predicts condition-mean effects rather than
within-condition single-cell distributions; downstream uses involving state
mixtures, rare subpopulations, or distributional distances will require
cell-level generators or transport models that preserve the same identity-held
source-token boundary.

\section{Conclusion}
\label{sec:conclusion}

Single-cell perturbation atlases often miss context--perturbation pairs even
when the same intervention has been measured in other contexts. In this paper,
we propose \proj, a source-response transport framework that predicts missing
recipient-context effects by mapping measured source responses through
train-only reliability-weighted source-to-recipient routes. On a
Perturb-CITE-seq melanoma immune-evasion evaluation, \proj reduces full-effect
MSE from $1.6490\times10^{-3}$ to $1.5809\times10^{-3}$ relative to a
recipient-local base predictor while remaining close to a stronger centralized
token-matched reference.

\bibliographystyle{plainnat}
\bibliography{references}

\end{document}